 \documentclass[pmlr,twocolumn,10pt]{jmlr} % W&CP article

\usepackage{booktabs}
\usepackage{siunitx}

\usepackage[switch]{lineno}

\usepackage{multirow}
\usepackage{graphicx} % Needed for \resizebox
\usepackage{booktabs} % For better lines
\usepackage{colortbl} % For row colors
\theorembodyfont{\upshape}
\theoremheaderfont{\scshape}
\theorempostheader{:}
\theoremsep{\newline}

\jmlrvolume{LEAVE UNSET}
\jmlryear{2024}
\jmlrsubmitted{LEAVE UNSET}
\jmlrpublished{LEAVE UNSET}
\jmlrworkshop{Machine Learning for Health (ML4H) 2024} % W&CP title

\title{Can large language models be privacy preserving and fair medical coders?}

\author{%
\Name{Ali Dadsetan}\Email{ali.dadsetan@dal.ca}\\
\Name{Dorsa Soleymani}\Email{dr474970@dal.ca}\\
\Name{Xijie Zeng}\Email{xijie.zeng@dal.ca}\\
\Name{Frank Rudzicz}\Email{frank@dal.ca}\\
\addr Dalhousie University, Canada; Vector Institute, Canada
\\
}

\begin{document}

\maketitle

\begin{abstract}
Protecting patient data privacy is a critical concern when deploying machine learning algorithms in healthcare. Differential privacy (DP) is a common method for preserving privacy in such settings and, in this work, we examine two key trade-offs in applying DP to the NLP task of medical coding (ICD classification). Regarding the privacy-utility trade-off, we observe a significant performance drop in the privacy preserving models, with more than a 40\% reduction in micro F1 scores on the top 50 labels in the MIMIC-III dataset. From the perspective of the privacy-fairness trade-off, we also observe an increase of over 3\% in the recall gap between male and female patients in the DP models. Further understanding these trade-offs will help towards the challenges of real-world deployment.
\end{abstract}

\paragraph*{Data and Code Availability} We use the MIMIC-III
% and MIMIC-IV
dataset, which is publicly available, under a specific training course and a data use agreement. We are making our code  publicly available.
\paragraph*{Institutional Review Board (IRB)} This work is exempt from IRB review, as it exclusively uses data that is publicly available.

\section{Introduction}
%Following the success of  language models in a variety of natural language processing (NLP) tasks \citep{llama2,gemma,phi},
Applying pre-trained, transformer-based models  has been a subject of recent interest in the medical domain \citep{huang-etal-2022-plm,yang-etal-2022-knowledge-injected, wang-multi-stage}. While these applications offer many benefits, some privacy concerns may exist including leaking sensitive information from  training datasets \citep{carlini2021extracting}. Moreover, if a curious actor gets access to the parameters of a model trained on sensitive data, they may be able to use methods to check if a specific piece of information has been part of the training dataset for that model \citep{carlini2022membership,mireshghallah-etal-2022-quantifying}. 

Privacy-preserving learning  has been applied effectively in NLP tasks outside of healthcare. Gradient-perturbation-based differential privacy \citep{abadi2016deep,li2022large} is a popular set of methods that have been applied successfully to language generation and sentence classification using LLMs \citep{li2022large}. In parallel, differential privacy has been studied in healthcare, which is considered to be challenging. \citet{suriyakumar_chasing_2020} showed that the performance gap between private and non-private algorithms is bigger in healthcare datasets, and hypothesized that the cause is long-tail distributions common in healthcare. They also showed a privacy/fairness trade-off in which privacy-preserving versions of algorithms diminished the influence of minority groups on inference. However, in the non-private versions, this was not the case.

One shortcoming of that study was that the authors only considered time-series and imaging data, and  not  text. While there is existing literature on privacy in text data, there remains a need for further exploration of privacy trade-offs specifically in NLP applications within the healthcare domain.
In this work, we  partially fill this gap by studying various trade-offs of applying differentially private learning algorithm to the task of assiging International Classification of Diseases (ICD) codes to discharge notes, which is a common application of NLP in healthcare.

\section{Background}

A popular mathematical formalism for the definition of private algorithms in machine learning is differential privacy
(DP). In a differentially private algorithm, inclusion or exclusion of a particular sample in the training
dataset has a limited effect on any expected outcomes \citep{dwork2014algorithmic}. More formally, we say that a randomized algorithm \(\mathcal{M}\) is \((\epsilon, \delta)\) approximately differentially private, if given
any two datasets differing only by one data point \(X, X'\), the probability that the output of the
algorithm on these datasets would occur in any specific event, such as \(Y\), would not differ more than a multiplicative factor of \(\exp(\epsilon)\) plus a constant \(\delta\):
\[
    \mathbb{P}(\mathcal{M}(X) \in Y) \leq \exp(\epsilon) \mathbb{P}(\mathcal{M}(X')\in Y) + \delta.
\]
In gradient-based learning, differential privacy of an algorithm can be guaranteed by modifying the gradient in a specific way before applying the optimization method of choice. Given a batch of \(B\) samples \(\set{B}\) and the corresponding gradients \(\nabla\mathcal{L}_i\), \(i\in\set{B}\), the DP-SGD algorithm \citep{abadi2016deep} works by clipping each \(\nabla\mathcal{L}_i\)  to ensure its length does not exceed \(C\), then obtaining the average of the clipped gradient for the batch, and finally adding a isotropic normal noise with covariance matrix \(\sigma^2\mathbb{I}\):
\[
\Tilde{\vec{g}} = \frac{1}{B}\sum_{i\in\set{B}}\text{Clip}(\nabla\mathcal{L}_i, C) + \mathcal{N}(0, \sigma^2\mathbb{I}).
\]
Instead of using the usual \(\vec{g}=\frac{1}{B}\sum_{i\in\set{B}}\nabla\mathcal{L}_i\),  \(\Tilde{\vec{g}}\) is  used by the optimization algorithm. The final ``privacy loss", \(\epsilon\),  depends on the magnitude of the added noise, \(\sigma\) ,the noise multiplier, the number of gradient updates applied to the model, the sampling rate, \(q=\frac{B}{N}\) where \(N\) is the dataset size, and the chosen approximation constant, \(\delta\). This relationship has been studied extensively \citep{abadi2016deep,mironov2019r,wang2019subsampled,koskela2020computing,gopi2021numerical}, and tight bounds exist \citep{gopi2021numerical}.

\subsection{Differential Privacy for LLMs}
Recent work explored applying differential privacy techniques to LLMs to enable private fine-tuning while maintaining both utility and privacy. In fact, large pre-trained language models (PLMs) can be effective differentially private learners when fine-tuned with appropriate hyperparameters and objectives \citep{li2022large}. Authors in \citep{li2022large} showed that training PLMs such as RoBERTa for sentence classification tasks (e.g., MNLI) and GPT-2 for language generation tasks (e.g., E2E) with strong privacy guarantees \(\epsilon \in\{3,8\}\) is achievable with only marginally reduced performance compared to their non-private counterparts.

Another challenge of applying gradient-perturbation-based DP algorithms to NLP tasks is their increased memory and computational complexity. From this perspective, the most demanding aspect of the algorithm is gradient clipping \citep{li2022large,bu2023differentially}. The clipping operation with clip factor \(C\) has the role of controlling the sensitivity of the gradient, and is defined as 
\[\mathrm{Clip}(\nabla \mathcal{L}_i)=\min(1, \frac{C}{\|\nabla\mathcal{L}_i\|_2})\nabla\mathcal{L}_i.\]
Clipping in DP-SGD occurs on a per-example basis. The na\"ive implementation may fail to leverage the parallel  capabilities of GPUs, significantly increasing execution time by disregarding batch computations and calculating gradients for each sample. Other approaches involve implicit clipping, as seen in methods such as \textit{ghost clipping} \citep{li2022large} and \textit{Book Keeping} \citep{bu2023differentially}. 

A third option is to perform clipping on a per-group basis, where a group may be defined as either a layer of the neural network or a set of parameters for a layer (e.g., weights of a linear layer) \citep{mcmahan2017learning,bu2023accuracy}. Here, to ensure that the gradient for each sample does not exceed a maximum length of  \(C\), we assign a clipping threshold to each  \textit{group} of parameters. Specifically, the parameter set \(\theta\) is partitioned into \(k\) groups, each representing parameters of a layer, denoted as \(\theta=(\theta_1,\dots,\theta_k)\). Correspondingly, the gradients are partitioned as \(\nabla_\theta \mathcal{L}=(\nabla_{\theta_1}\mathcal{L},\dots,\nabla_{\theta_k}\mathcal{L})\).  To ensure that the overall gradient \(\nabla_\theta \mathcal{L}\) does not exceed the maximum length \(C\), we clip each group’s gradient \(\nabla_{\theta_i}\mathcal{L}\) such that its length is limited to \(\frac{C}{\sqrt{k}}\).

If we choose groups to correspond to the parameters of each layer, the computational overhead for this clipping method will be significantly lower compared to other methods. Despite its simplicity, this performs relatively well for LLMs, exhibiting only a marginal performance gap compared to the conventional all-layer clipping method \citep{bu2023accuracy}.

\subsection{Privacy-Fairness trade-off in healthcare}
\citet{suriyakumar_chasing_2020} studied various trade-offs of differential privacy  in health care settings which they showed was more challenging than in other domains, and they argued that this is because of the long tails presented in the distribution of health care data. 

They showed that differential privacy unfairly changes the group influence of black vs white patients. In non-private training, the most influential data points for the model's decisions regarding Black patients were those from Black patients. However, in private training, data from White patients had the greatest influence on the model's predictions for Black patients. They showed this across varying privacy budgets, and for image and time-series domains. However, this  privacy-fairness trade-off was not applied to NLP tasks in healthcare.

\subsection{ICD classification}

ICD coding is a multi-label classification task of assigning ICD codes using  patient data, such as discharge summaries \citep{vu2020label}. Automatic ICD coding from clinical text has been an active area of research, with significant advancements in deep learning approaches \citep{mullenbach2018-explainable, Xie2019, li2019icdcodingclinicaltext, vu2020label, huang-etal-2022-plm}.

\paragraph{LAAT with PLM-ICD}  The label attention model (LAAT)  \citep{vu2020label} uses a Bi-LSTM encoder and label-specific attention, which we refer to as LA\textsubscript{LAAT}, to create label-aware document representations. This mechanism transforms a sequence of hidden representations $H\in\mathbb{R}^{d_h \times N}$, where \(N\) is the length of the sequence, into label-specific representations $V \in \mathbb{R}^{d_h \times L}$, where $L$ is the number of unique medical codes. Attention is computed as:
\[
A = \text{softmax}(UZ), \text{ where } Z = \tanh(PH)
\]
Here, $U \in \mathbb{R}^{L \times d_p}$ and $P \in \mathbb{R}^{d_p \times d_h}$ are learnable matrices, and $d_p$ is the dimension of the intermediate representation of the document. Then, the label specific representations, \(V\), can be computed as \(HA^T\). Assignment of labels happens by feeding these label specific representations  into  linear classifiers specific to each label with sigmoid activation. This approach demonstrated superior performance on the multi-label classification tasks on ICD code assignment, on the MIMIC-III dataset. Specifically, LAAT achieved a micro-AUC of 98.8, micro-F1 score of 57.5 and a macro-F1 score of 9.9 on the full MIMIC-III dataset.

More recently, \citet{huang-etal-2022-plm} introduced PLM-ICD, which used PLMs as the encoder for ICD coding. To handle clinical notes that exceed the maximum input length of PLMs, the authors implemented a method which divides long documents into multiple segments, processes each segment independently through the PLM.

\section{Methodology}
\subsection{Dataset}
We use discharge summaries from  MIMIC-III  
\citep{johnson2016mimic},

 which contain de-identified health records from hospital admissions. We focus on classifying the top 50 most frequent ICD codes.
 We prepared the data following the method outlined by \citet{edinAutomatedMedicalCoding2023}, which builds on the approach proposed by \citet{mullenbach2018-explainable} 
 and reduced the number of records to facilitate a direct comparison with previous studies  \citep{mullenbach2018-explainable}. We filtered the dataset to include only instances that contained at least one of the top 50 most frequent codes and excluded patients who did not have any of these codes.

 The dataset was split into training, validation, and test sets with a 70/15/15 ratio, as per \citet{mullenbach2018-explainable}. 
 between the training and evaluation phases. In the final dataset, we have 8066, 1729, and 1573 admissions in the training, test, and validation datasets, respectively. These admissions were related to 7501, 1526, and 1329 patients, respectively. For more details about pre-processing, please refer to \appendixref{app:pre}.

\subsection{Model Architecture}
For the task of ICD-classification, we  follow the PLM-ICD architecture \citep{huang-etal-2022-plm}, which itself uses the label attention (LAAT) module \citep{vu2020label}. 

For the PLM component, we have used  \textit{Meditron-7b} , \citep{chen2023meditron} and \textit{Gatortron-large} \citep{yang2022large} which are both language models trained specifically for the medical domain. The loss function for our model is the binary cross entropy loss between the predicted probabilities from the model and the true labels. 

\subsection{Training and evaluation}
For both of the privacy-preserving and regular versions of our models, we use the AdamW optimizer with a learning rate of \(5\times 10^{-5}\), weight decay of \(0.1\), $\beta$ parameters of \((0.9, 0.95)\), $\epsilon$ of \(10^{-5}\), and a linear decay learning rate schedule with a 200-step  warm-up. The batch size was set to 32, and we ran our training on 4 A100 80GB Nvidia GPUs. We use the Hugging Face Transformers library \citep{wolf2020huggingfacestransformersstateoftheartnatural} for model implementation and the Deepspeed \citep{Deepspeed2020} and PyTorch 
for distributed training.

\paragraph{Non-Private Training}
On MIMIC-III dataset, we trained our model for 5 epochs. We monitored model loss and f1-score on validation dataset during training, both reaching their optimum on second epoch. This was sufficient to achieve results very close to state of the art on the top 50 labels.

\paragraph{Private Training}
In the private setting, our models were trained for a maximum of 20 epochs. In our experiments, training for more epochs with the same noise level did not improve the loss and micro F1 score on the validation set. Our implementation of gradient perturbation combines two key techniques: ghost clipping and group clipping. Ghost clipping, as described by \citet{li2022large}, allows for efficient computation of gradient norms with only materializing per-example gradients of single layer, and also computing the per example gradients of the embedding layers in a more efficint way. We use this technique alongside with group clipping \citep{bu2023accuracy} , where we define parameter groups as the weights and biases of all the linear layers of the models, and the weights for root mean square normalization components in Meditron, and the layer norms in Gatortron. We use a noise multiplier of \(0.05\) and clipping constant \(0.1\). The clipping constant of \(0.1\) was found to be optimal for other tasks \citep{li2022large}, and we choose the noise multiplier in a way that the final privacy loss will be less than 10. In our implementation, we largely followed the Pytorch Extending approach in the \textit{fastDP} library \citep{bu2023differentially}.

To compute the \(\epsilon\), the privacy loss, we employ the Privacy Loss Random Variable (PRV) accountant. This accountant computes the privacy loss based on the magnitude of the added noise, the number of updates, and the sampling rate \citep{gopi2021numerical}. This provides tighter privacy bounds, allowing for more accurate privacy guarantees. Setting the value of \(\delta\) to the inverse of the number of samples in the training dataset, the upper bound for privacy loss was \(9.97\), which is looser than what was used in previous work in non-healthcare settings \citep{li2022large}.

For measuring fairness, we  measure the performance gap between different control groups (e.g., ethnicity, gender), as measured by the micro versions of metrics like parity, recall, and F1 score. 

\section{Results and Discussion}
For non-privacy-preserving versions, the micro F1 scores on the test set were 73.6\% (Gatortron) and 67.3\% (Meditron). Interestingly, the former is better than the performance previously reported for the same dataset and task \citep{medical_coding_review}. 
In the privacy-preserving versions, the micro F1 scores dropped significantly, to 27.2\% (Gatortron) and 30.8\% (Meditron). 
Compared with the smaller performance gaps reported by \cite{li2022large} and \cite{bu2023accuracy} on tasks such as language generation and sentence classification with metrics like BLEU and accuracy, our results reinforce those of \cite{suriyakumar_chasing_2020} that differential privacy in healthcare remains challenging. 

On the fairness side, we saw an increase in the gender gap in  model performance between privacy-preserving and non-privacy preserving models. For example, the gender recall gap with Meditron increased from 0.4\% to 3.6\%, and from 0.0\% to 3.4\% with Gatortron. This also reinforces the findings in other modalities, that privacy may affect protected groups unfairly \citep{suriyakumar_chasing_2020}. However, the effect of privacy on the performance gap between ethnic groups is model-dependent. For more details, please refer to \appendixref{apd:fairness}. 

Although ensuring patient privacy is crucial, our results highlight the challenging nature of applying differential privacy in  healthcare. In particular, these results call for further research in improving the privacy-fairness trade off in healthcare, so that optimizing one doesn't come at the cost of the other.

\acks % commenting for annonymization 
Resources used in preparing this research were provided, in part, by the Province of Ontario, the Government of Canada through CIFAR, and companies sponsoring the Vector Institute \url{https://vectorinstitute.ai/partnerships/current-partners/}.
% Xindy, Adil (maybe VectorLM seperately?), Vinith.  , vector sponsers?
\bibliography{jmlr-sample}

\appendix
\section{Detailed fairness comparison}\label{apd:fairness}
For gender groups, the results in \tableref{tab:gender-metrics} indicate a significant performance drop when differential privacy (DP) is applied . In the non-private setting, F1 scores for both male and female groups are similar, with Gatortron achieving 74\% and Meditron scoring 68\% for males, with females trailing by only 1\%. However, under the DP setting, F1 scores drop dramatically to around 29\% for males and 25\% for females with Gatortron, and 32\% for males and 29\% for females with Meditron. This demonstrates an increase in the gender gap in F1 scores, from 1\% in the non-private case to 4\% and 3\% in the DP versions of Gatortron and Meditron, respectively \figureref{fig:gender_gap}. Recall metrics follow a similar trend, with non-DP models performing significantly better in terms of gender fairness.

For ethnicity groups, the performance metrics are shown in \tableref{tab:ethnicity-metrics}. In non-DP models, Gatortron achieves an F1 score of 74\% for most ethnic groups, while Meditron scores range between 63\% and 68\%. In the DP setting, F1 scores drop to 29\% for White patients, 25\% for Black and Hispanic patients, and 22\% for Asian patients with Gatortron. Meditron’s DP performance shows F1 scores of 30\% for White and Asian groups, 31\% for Black patients, and 32\% for Hispanic patients. While the non-private version of Gatortron shows better fairness across ethnic groups than its DP version, Meditron's DP version appears to be more equitable across ethnicities compared to its non-DP version \figureref{fig:ethnicity_gap}.

\begin{figure}[tbp]
\floatconts
  {fig:model_performance}
  {\caption{Comparison of F1 score between models trained with DP and models trained regularly. There is a clear decrease in performance of the privacy preserving models.}}
  {\includegraphics[width=0.8\linewidth]{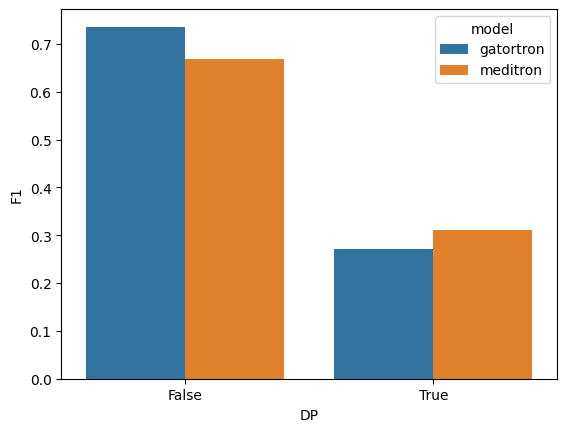}}
\end{figure}

\begin{figure}[tbp]
\floatconts
  {fig:gender_gap}
  {\caption{Comparison of gender gap in F1 score between models trained with DP and models trained regularly. The gender gap is clearly increased in privacy preserving models.}}
  {\includegraphics[width=0.8\linewidth]{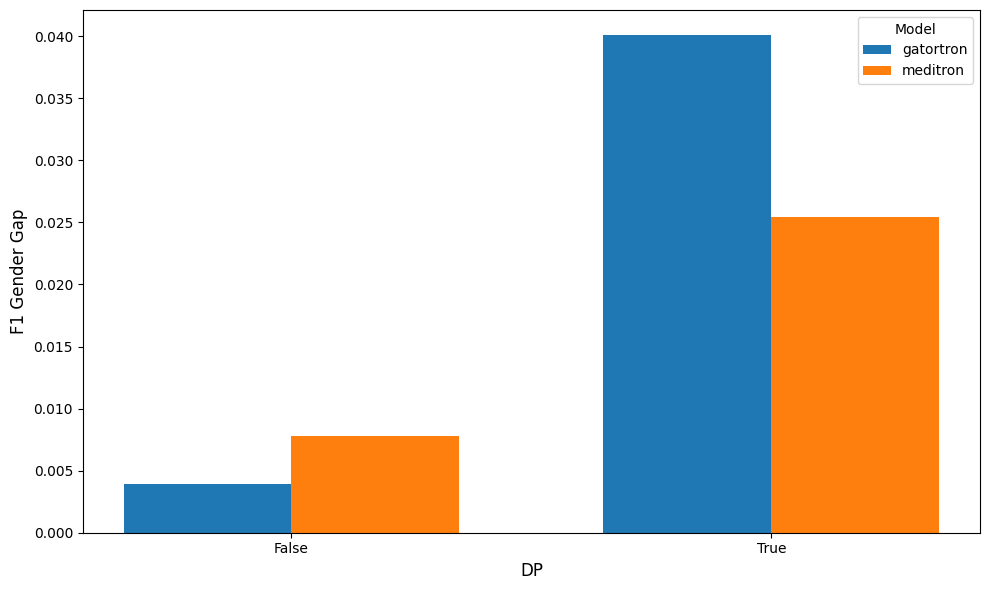}}
\end{figure}

\begin{figure}[tbp]
\floatconts
  {fig:ethnicity_gap}
  {\caption{Comparison of ethnicity gap, between white and black ethnicity, in F1 score between models trained with DP and models trained regularly. The ethnicity gap is increased in gatortron, and decreased in meditron. Similar results hold for other ethnicity pairs.}}
  {\includegraphics[width=0.8\linewidth]{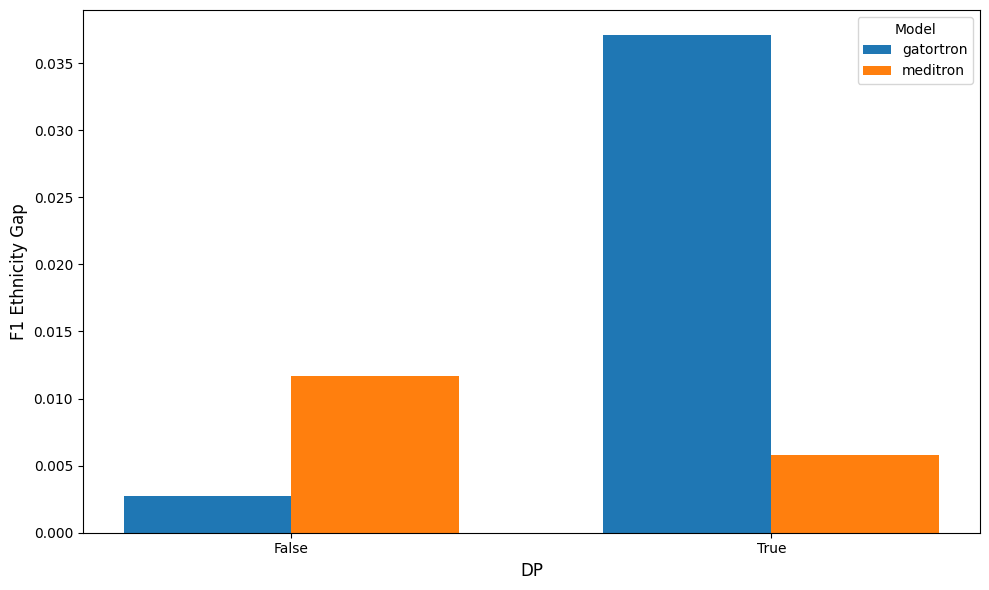}}
\end{figure}

\begin{table*}[tbp]
\floatconts
  {tab:gender-metrics}%
  {\caption{Performance metrics for gender groups}}%
  {\begin{tabular}{lccccc}
\toprule
\textbf{PLM} & \textbf{DP} & \textbf{Group} & \textbf{F1} & \textbf{Parity} & \textbf{Recall} \\
\midrule
\multirow{4}{*}{\textbf{Gatotron}} & \multirow{2}{*}{True} & Male & 0.29 & 0.05 & 0.20 \\
 & & Female & 0.25 & 0.04 & 0.16 \\
\cmidrule{2-6}
 & \multirow{2}{*}{False} & Male & 0.74 & 0.12 & 0.72 \\
 & & Female & 0.73 & 0.11 & 0.72 \\
\midrule
\multirow{4}{*}{\textbf{Meditron}}
& \multirow{2}{*}{True} & Male & 0.32 & 0.17 & 0.37 \\
 & & Female & 0.29 & 0.16 & 0.34 \\
\cmidrule{2-6}
 & \multirow{2}{*}{False} & Male & 0.68 & 0.13 & 0.68 \\
 & & Female & 0.67 & 0.12 & 0.67 \\
\bottomrule
\end{tabular}}
\end{table*}

\begin{table*}[tbp]
\floatconts
  {tab:ethnicity-metrics}%
  {\caption{Performance metrics for ethnicity groups}}%
  {\begin{tabular}{llccccc}
\toprule
\textbf{PLM} & \textbf{DP} & \textbf{Group} & \textbf{F1} & \textbf{Parity} & \textbf{Recall} \\
\midrule
\multirow{8}{*}{\textbf{Gatotron}} & \multirow{4}{*}{True} & White & 0.29 & 0.05 & 0.20 \\
 & & Black & 0.25 & 0.04 & 0.17 \\
 & & Hispanic & 0.25 & 0.04 & 0.17 \\
 & & Asian & 0.22 & 0.04 & 0.15 \\
\cmidrule{2-6}
 & \multirow{4}{*}{False} & White & 0.74 & 0.12 & 0.73 \\
 & & Black & 0.74 & 0.11 & 0.72 \\
 & & Hispanic & 0.72 & 0.11 & 0.71 \\
 & & Asian & 0.74 & 0.11 & 0.77 \\
\midrule
\multirow{8}{*}{\textbf{Meditron}} 
 & \multirow{4}{*}{True} & White & 0.30 & 0.17 & 0.35 \\
 & & Black & 0.31 & 0.15 & 0.35 \\
 & & Hispanic & 0.32 & 0.14 & 0.36 \\
 & & Asian & 0.30 & 0.15 & 0.37 \\
\cmidrule{2-6}
& \multirow{4}{*}{False} & White & 0.67 & 0.13 & 0.68 \\
 & & Black & 0.68 & 0.12 & 0.69 \\
 & & Hispanic & 0.68 & 0.10 & 0.64 \\
 & & Asian & 0.63 & 0.12 & 0.67 \\
\bottomrule
\end{tabular}}
\end{table*}

\section{Pre-processing}
\label{app:pre}
These are the preprocessing steps we performed to prepare our dataset for model input.
\begin{enumerate}
\item Converted all text to lowercase.
\item Replacing special characters with a space (" ").
\item Removed standalone numbers; for example, '120mg' was retained, while '700' was excluded.
\item Removed extra whitespace to ensure clean text.
\item Removed rows that did not contain ICD9 codes.
\item Eliminated duplicate ICD9 codes within each record.
\item Filtered the dataset to keep only the top 50 most frequent ICD9 codes.
\item Deleted rows that did not contain any of the top 50 ICD9 codes.
\item Split the dataset into training, testing, and validation sets with a 70/15/15 ratio.
\end{enumerate}

You can find a flow chart in \figureref{fig:preprocessing}.

\begin{figure}[htbp]
\floatconts
  {fig:preprocessing}
  {\caption{Pre-processing flow chart.}}
  {\includegraphics[width=0.5\linewidth]{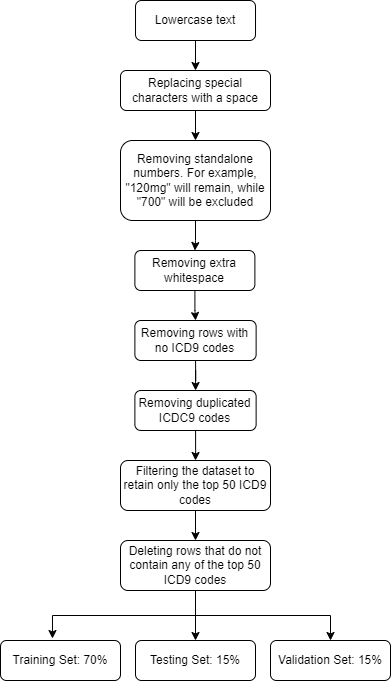}}
\end{figure}

\end{document}